\documentclass[conference]{IEEEtran}
\IEEEoverridecommandlockouts
\usepackage{cite}
\usepackage{amsmath,amssymb,amsfonts}
\usepackage{graphicx}
\usepackage{textcomp}
\usepackage{xcolor}
\usepackage{pifont}
\usepackage[hidelinks]{hyperref}
\def\BibTeX{{\rm B\kern-.05em{\sc i\kern-.025em b}\kern-.08em
    T\kern-.1667em\lower.7ex\hbox{E}\kern-.125emX}}

\usepackage{booktabs}      
\usepackage{array}
\usepackage{multirow}       
\usepackage{url}
\usepackage{balance}
\usepackage{microtype}
\usepackage{xspace}        
\usepackage{algorithm}      
\usepackage{algpseudocode}  %
\graphicspath{{figures/}}
\definecolor{tabred}{HTML}{D62728}
\graphicspath{{figures/}}

\newcommand{\headlinename}{\textsc{ATEST}}
\newcommand{\sys}{\textsc{\headlinename}\xspace}   
\newcommand{\atcal}{\textsc{\headlinename}-Calibrated\xspace}    
\newcommand{\atagt}{\textsc{\headlinename}-Online\xspace}       
\newcommand{\fixed}{Fixed-4\xspace}                
\newcommand{\adcal}{Adaptive-Cal\xspace}           
\newcommand{\Eocc}{E_{\mathrm{occ}}}            \usepackage{colortbl}                    
\definecolor{oursbg}{HTML}{E4EFF6}       
\newcommand{\ours}{\cellcolor{oursbg}}   

\newcommand{\cmark}{\ding{51}}   
\newcommand{\xmark}{\ding{55}}   

\begin{document}
\title{\headlinename: Adaptive Resource-Aware Test-Time Scaling for LLM Inference}

\author{
\IEEEauthorblockN{Mobina Kashaniyan and Ali Jannesari}
\IEEEauthorblockA{
Department of Computer Science\\
Iowa State University, USA \\
\{mobina, jannesar\}@iastate.edu
}
}

\maketitle

\begin{abstract}
Test-time scaling (TTS) improves the reasoning of large language models (LLMs) by generating several samples for a query and voting on the answer.
However, a fixed sampling budget ignores two facts: different queries need different amounts of reasoning, and the cost of one more sample changes with GPU load.
In our measurements, the same four-sample query uses up to 4.1$\times$ more energy when other jobs load the GPU.
Existing adaptive methods mainly set the budget from answer agreement, question structure, or question difficulty.
They decide how much reasoning a query needs, but not what that reasoning costs on the GPU right now.
We present \sys, a runtime framework for resource-aware TTS with two modes.
Its default mode, \atagt, needs no offline calibration: it starts with one sample, checks whether the plurality answer is stable, reads the live GPU state, predicts the cost of the next step with an online cost model, and decides whether to stop or continue.
For stable deployments that can reuse a one-time profile, \sys also offers an offline mode, \atcal, which picks the budget from accuracy and cost models built offline.
We evaluate \sys with Phi-3-mini on an NVIDIA A100 across four reasoning and question-answering datasets under a controlled idle-to-high GPU load schedule, and repeat the study on an NVIDIA L40S and with a second model, Qwen2.5-3B.
On the A100, \atagt answers every query, never exceeds the energy budget, gives the highest or statistically tied accuracy on all four datasets, and uses fewer samples as load rises.
The same pattern holds on the L40S and with Qwen2.5-3B, where \atagt again answers every query within the budget.
On GSM8K at high load, it keeps 60.0\% accuracy, while \atcal drops to 44.0\% because it defers 42.4\% of the queries.
A controlled ablation shows that online cost learning is what makes the budget follow the load. Including offline calibration, \atagt uses 42$\times$ less GPU time and 48$\times$ less energy than \atcal.
Its main costs are higher tail latency and lower throughput on short tasks, and offline calibration remains strong only when the load is stable.
These results show that TTS on shared GPUs is a runtime resource-management problem, not only an accuracy problem.
\end{abstract}


\begin{figure}[t]
    \centering
    \includegraphics[width=\columnwidth]{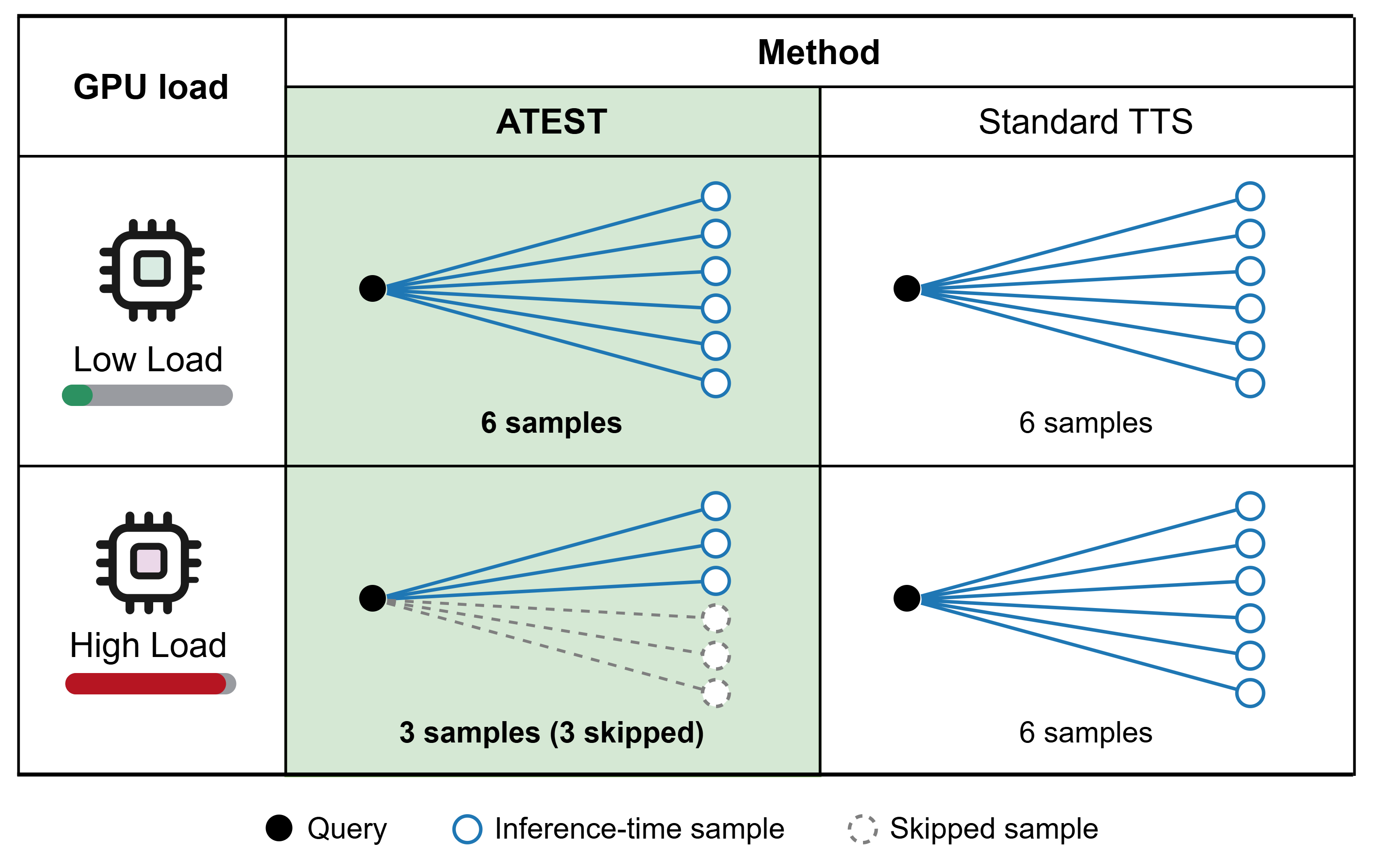}
    \caption{The idea of \sys. Standard test-time scaling (TTS) generates the same number of samples for a query whether the GPU is idle or busy. \sys  looks at the live GPU load. When the GPU is busy and each sample costs more energy, it skips samples that are not worth their cost.}
    \label{fig:teaser}
\end{figure}

\begin{figure*}[t]
    \centering
    \includegraphics[width=0.97\textwidth]{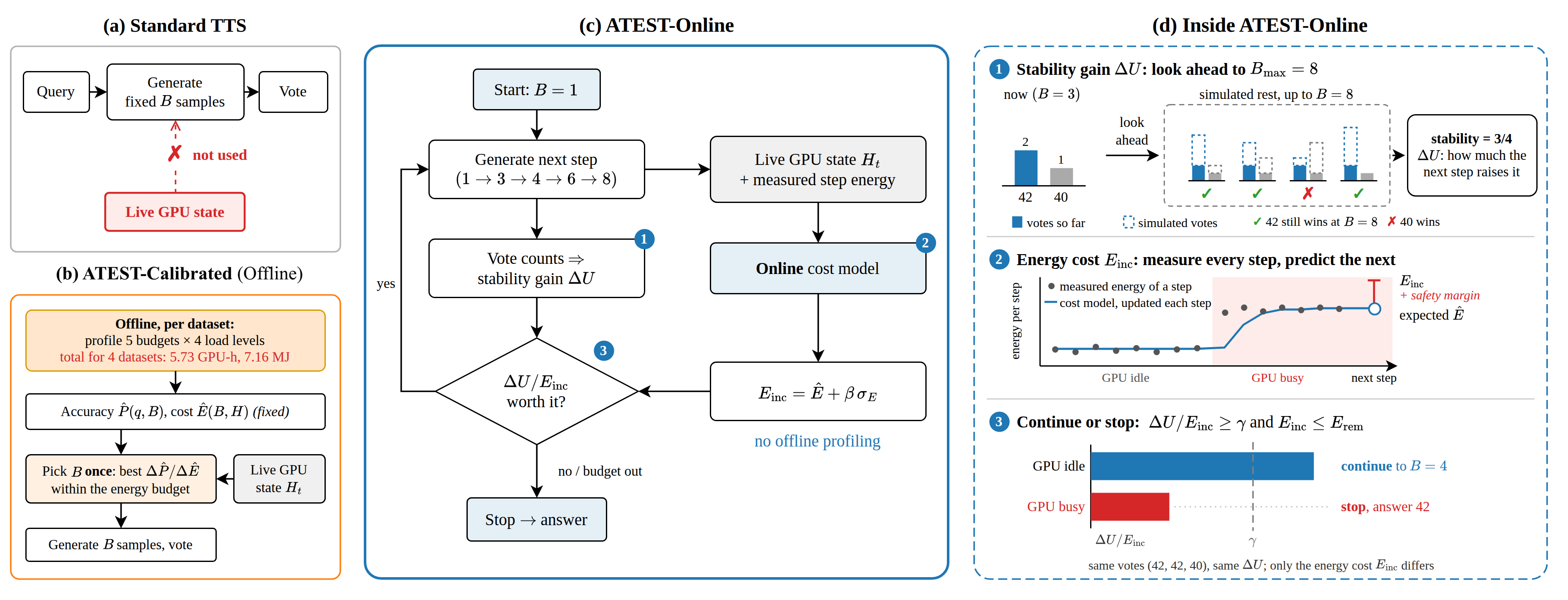}
\caption{Overview of \sys. (a)~Standard TTS generates a fixed number of samples $B$ and does not use the live GPU state. (b)~\atcal, the offline mode of \sys, builds accuracy and cost models offline for each dataset and picks $B$ once per query. (c)~\atagt starts with one sample and adds samples step by step ($1\!\to\!3\!\to\!4\!\to\!6\!\to\!8$). (d)~An illustration of one decision: \textcircled{\scriptsize 1}~a Monte Carlo look-ahead to $B_{\max}=8$ estimates how likely the current winner is to stay the winner, and $\Delta U$ is how much the next step raises this; \textcircled{\scriptsize 2}~the online cost model learns from the measured energy of every step, follows the load, and predicts the next step's energy with a safety margin; \textcircled{\scriptsize 3}~with the same votes, \atagt continues on an idle GPU but stops on a busy one.}
    \label{fig:overview}
\end{figure*}

\section{Introduction}
\label{sec:introduction}

Large language models (LLMs) give better answers when they use more computation at inference time.
A simple and popular way to do this is \emph{test-time scaling} (TTS) with voting: the model generates several answers (samples) for the same question, and the system returns the most common one~\cite{wei2022chain,wang2023selfconsistency}.
More samples usually mean better answers, but they also mean more GPU work, more energy, and longer waiting time.

Because of this cost, recent work makes TTS \emph{adaptive}.
These methods give more samples to hard queries and fewer to easy ones, using signals such as answer agreement, query difficulty, or reasoning progress~\cite{aggarwal2023adaptive,li2024esc,snell2024scaling,wan2025rasc,wang2025dsc,kashaniyan2026adaptive}.
They answer one question well: \emph{how many samples does this query need?}
However, they treat the cost of one sample as a fixed number. On a real system, this cost is not fixed.
LLM performance depends on batching, scheduling, and memory management~\cite{yu2022orca,kwon2023vllm}, and the same number of samples can have very different energy, latency, and throughput costs~\cite{kashaniyan2026samplecount}.

Many GPUs in clusters and HPC centers are \emph{shared}: other jobs run on the same GPU at the same time and compete with the LLM for compute, memory bandwidth, and power, and this competition changes over time.
In our measurements on an A100, the same four-sample query uses up to 4.1$\times$ more energy when the GPU is heavily loaded (Fig.~\ref{fig:motivation}).
For example, a GSM8K query costs 121.7~J on an idle GPU and 455.6~J on a busy one.
A controller that ignores this change either wastes energy when the GPU is busy, or uses up its energy budget too early and must \emph{defer} later queries, which means refusing to answer them.

So a TTS controller on a shared GPU must answer two questions at the same time:
\begin{enumerate}
    \item \emph{Is one more sample useful for this query?}
    \item \emph{What does that sample cost on the GPU right now?}
\end{enumerate}

We present \textbf{\sys}, a resource-aware TTS framework that answers both questions.
Fig.~\ref{fig:teaser} shows the idea: when the GPU is busy, \sys skips samples that are not worth their higher cost.
\sys has two modes that share the same budget ladder, GPU monitor, and energy accounting (Fig.~\ref{fig:overview}).
Its default mode, \textbf{\atagt}, needs no offline profiling.
It works as a closed control loop: it generates samples in small steps, checks whether the vote is becoming stable, reads the live GPU state, measures what each step really cost, updates an online cost model, and stops as soon as another step is not worth its cost. 

Some deployments, however, run the same model and workload on a GPU whose load rarely changes, or need low tail latency.
For these cases \sys also offers an offline mode, \textbf{\atcal}, which profiles each dataset once and picks the whole budget per query from accuracy and cost models, together with the live GPU state.
Choosing $B$ once lets it generate all samples in one batch, which gives lower tail latency and higher throughput on short tasks.
The price is the profile itself, which took 5.73 GPU-hours and 7.16~MJ for our four datasets, and a cost model that does not follow the GPU when the load changes.
Including profiling, \atagt uses 42$\times$ less GPU time and 48$\times$ less energy than \atcal on the same 2,000 queries.

This difference matters most when the GPU becomes busy.
On GSM8K under high load, \atcal must defer 42.4\% of the queries, so it answers only 44.0\% of them correctly.
Under the same load, \atagt answers every query, keeps 60.0\% accuracy, and uses less energy per query (175.8~J vs.\ 206.4~J).
Across all four datasets, \atagt never defers a query, never exceeds the energy budget, and gives the highest or tied-best accuracy. We also test \emph{why} \atagt changes its budget.
In a controlled ablation, we keep the remaining-energy signal fixed, so that budget depletion cannot drive the decision.
Without online cost learning, the controller uses the same mean budget ($B=3.12$) at every load level on all four datasets.
With online cost learning, the budget follows the load.
So the adaptation comes from the runtime resource signals, not from the budget running out.

\smallskip
\noindent\textbf{Contributions.} 
\noindent This paper makes three contributions:
\begin{enumerate}
    \item We show with measurements that the cost of a TTS sample is not fixed: on a GPU shared with other jobs, the same four-sample query uses up to 4.1$\times$ more energy under high load. Based on this, we treat adaptive TTS as a runtime resource-management problem, where each decision weighs the value of more reasoning against its current cost.
    \item We introduce \sys with two modes. Its default mode, \atagt, learns the cost of sampling online and decides step by step, with no dataset-specific profiling. Its offline mode, \atcal, picks the budget once from offline accuracy and cost models together with the live GPU state, for deployments that can reuse a one-time profile.
    \item We show when each mode should be used. Under changing load, \atagt answers every query and stays within the energy budget, while \atcal defers up to 42.4\% of the queries. Under stable load, offline calibration remains a strong choice. A controlled ablation shows that online cost learning is what makes the budget follow the load.
\end{enumerate}

\section{Background and Related Work}
\label{sec:background}

\subsection{Test-Time Scaling and Adaptive Sampling}

Test-time scaling (TTS) uses more computation at inference time to get better answers \cite{muennighoff-etal-2025-s1}.
A common form is to generate several answers for the same query and combine them.
Chain-of-thought prompting~\cite{wei2022chain}, \cite{chen2026towards},\cite{chen2025reasoning} lets the model write out its reasoning, and self-consistency~\cite{wang2023selfconsistency,taubenfeld2025confidence} samples several reasoning paths and returns the answer that appears most often.
We call the number of samples the \emph{sampling budget}~$B$. The same budget is not useful for every query.
Easy questions are often answered correctly after one or two samples, while hard questions may need more.
Adaptive methods therefore change $B$ per query.
Adaptive-Consistency~\cite{aggarwal2023adaptive},\cite{huang2025efficient} and Early-Stopping Self-Consistency~\cite{li2024esc}, \cite{wan2024dynamic}, \cite{aghazadeh2025cges,liu2025answer,sun2026stop} stop sampling once the answers agree.
Other methods use question difficulty, answer confidence, or reasoning progress~\cite{snell2024scaling,wan2025rasc,wang2025dsc,kashaniyan2026adaptive}.

These methods remove many unnecessary samples.
However, they decide only from the reasoning side: they ask whether more samples would change the answer, but not what those samples cost on the GPU right now.
Recent work shows that the number of samples alone does not describe the systems cost of TTS, because the same samples can have very different energy, latency, and throughput costs depending on how they run~\cite{kashaniyan2026samplecount}.
\sys adds this missing systems side to adaptive TTS.

\subsection{GPU-Aware and Energy-Efficient LLM Serving}

LLM serving systems\cite{pan2025survey,park2025survey} show that runtime conditions strongly affect inference performance.
Orca~\cite{yu2022orca} and vLLM~\cite{kwon2023vllm} improve batching and GPU memory management.
Llumnix~\cite{sun2024llumnix} moves requests between model instances at runtime to balance load and reduce latency, and Bullet~\cite{lin2026bullet} changes GPU resource allocation during execution based on current performance conditions. Energy is also a main goal in LLM serving.
Samsi et al.~\cite{samsi2023words} measured the energy cost of LLM inference on V100 and A100 GPUs.
POLCA~\cite{patel2024polca} studies power management and power oversubscription in LLM clusters, and DynamoLLM~\cite{stojkovic2025dynamollm} reconfigures inference clusters to save energy while meeting latency targets.
throttLL'eM~\cite{kakolyris2025throttle} adjusts GPU frequency and resource configuration to reduce inference energy while meeting latency targets.
EnerTune~\cite{sinha2026beyond} studies energy-aware GPU sharing, models how GPU allocation, frequency, and batch size affect power and performance, and shows that large profiling spaces can make offline profiling expensive.

These systems decide \emph{how} to run a given amount of work: where to place requests, how to batch them, and how to set the hardware.
They do not decide \emph{how much} reasoning work each query should get.
\sys makes this second decision, so it complements these systems and can run on top of them.

\subsection{Research Gap}
\begin{table}[t]
\centering
\caption{Families of TTS methods. Indented rows are the methods we run.}
\label{tab:related}
\footnotesize
\setlength{\tabcolsep}{4pt}
\begin{tabular}{@{}lcc@{}}
\toprule
\textbf{Method} & \textbf{Cost unit} & \textbf{Live GPU state} \\
\midrule
\textbf{Self-consistency family}~\cite{wang2023selfconsistency,aggarwal2023adaptive,li2024esc,wan2025rasc} & samples & \xmark \\
\quad $\hookrightarrow$ \fixed (baseline) & samples & \xmark \\
\midrule
\textbf{Adaptive allocation family}~\cite{wang2025dsc,snell2024scaling,kashaniyan2026adaptive} & tokens, compute & \xmark \\
\quad $\hookrightarrow$ \adcal (baseline) & energy (model) & \xmark \\
\midrule
\textbf{\sys (this work)} & & \\
\quad $\hookrightarrow$ \atcal (offline mode) & energy (model) & \cmark \\
\ours\quad $\hookrightarrow$ \atagt (default) & \ours energy (live) & \ours\cmark \\
\bottomrule
\end{tabular}
\end{table}
Adaptive TTS decides how much reasoning a query needs but treats the cost of a sample as fixed, while inference and energy systems manage the GPU but do not change how much reasoning each query gets.
On a shared GPU, these decisions depend on each other: one more sample is worth it only if the answers so far and the current GPU cost justify it.
No prior method in Table~\ref{tab:related} reads the live GPU state; the closest, \adcal~\cite{kashaniyan2026adaptive}, uses offline-calibrated models.
\sys makes this decision at runtime using both signals, and \atagt learns the cost online without per-dataset profiling.

\section{Motivation}
\label{sec:motivation}

\begin{figure}[t]
    \centering
    \includegraphics[width=\columnwidth]{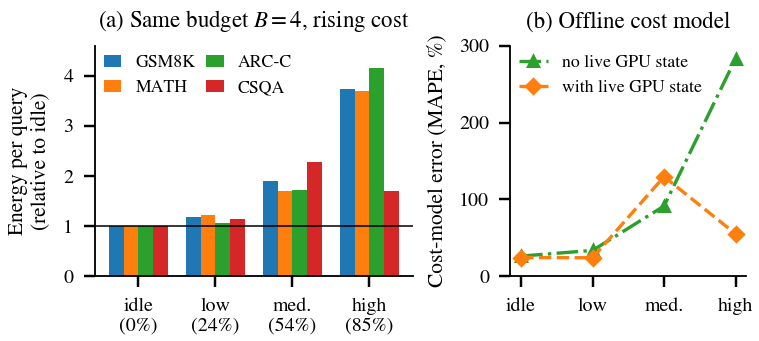}
    \caption{Motivation. (a)~Energy per answered query for a fixed budget of $B=4$ at four GPU load levels, relative to idle. The value under each level is the measured GPU utilization. (b)~Error of the offline cost model, averaged over four datasets. Without live GPU state (\adcal), the error grows from 26\% at idle to 283\% at high load. Adding live GPU state (\atcal) lowers the high-load error to 55\%, but the error is still large at medium load (130\%).}
    \label{fig:motivation}
\end{figure}

When several jobs share one GPU, they compete for the same compute units, memory bandwidth, and power limit.
As a result, the same microbatch runs longer, and the GPU draws more power while it runs.
In our setup, each microbatch is charged for the GPU energy used while it occupies the GPU (\S\ref{sec:methodology}), so the same number of samples costs more joules when the GPU is busy.
This cost depends on the current load, which changes over time and is not known in advance.
Fig.~\ref{fig:motivation} shows two observations from our experiments.

\noindent\textbf{Observation 1: the same budget can cost several times more.}
With a fixed budget of $B=4$, the energy per answered query rises with load on all four datasets.
At high load it is 3.7$\times$ the idle value on GSM8K and MATH, and 4.1$\times$ on ARC-Challenge.

\noindent\textbf{Observation 2: an offline cost model does not follow the load.}
A cost model fitted offline predicts the cost of a budget before it runs.
Fig.~\ref{fig:motivation}(b) shows that its prediction error grows quickly as the GPU becomes busy.
Adding the live GPU state as an input helps, but it does not remove the problem.

These observations lead to four design goals for a TTS controller on a shared GPU:
\begin{itemize}
    \item \textbf{G1 -- Spend samples where they help.} Give more samples only to queries whose answer is still uncertain.
    \item \textbf{G2 -- Follow the live cost.} Use the cost of a sample \emph{now}, not the cost measured during profiling.
    \item \textbf{G3 -- Avoid expensive profiling.} Do not require a new offline calibration for every dataset.
    \item \textbf{G4 -- Stay safe and serve everyone.} Never exceed the energy budget, and avoid deferring queries.
\end{itemize}
\atcal meets G1, partly meets G2, and does not meet G3.
\atagt is designed to meet all four.

\section{Problem Formulation}
\label{sec:problem}

We serve a stream of $N$ queries $q_1,\dots,q_N$ on one GPU that is shared with other jobs.
For each query $q_i$, the controller chooses a budget $B_i\in\mathcal{B}=\{1,3,4,6,8\}$.
The GPU condition at time $t$ is $H_t$.
Because $H_t$ changes, the same budget can have a different energy cost at different times.

All queries share one energy budget $E_{\max}$.
Let $E_i$ be the measured GPU energy used for query $q_i$.
The total must stay within the budget:
\begin{equation}
\sum_{i=1}^{N} E_i \le E_{\max}.
\label{eq:budget}
\end{equation}
A query is admitted only if its predicted cost fits its share of the remaining budget, $E_{\mathrm{rem}}/N_{\mathrm{rem}}$, where $N_{\mathrm{rem}}$ is the number of queries left; otherwise it is \emph{deferred}.
A deferred query gets no answer, and we write $B_i=0$.
The \emph{defer rate} is the fraction of deferred queries, and the \emph{service coverage} is one minus the defer rate. The goal is to answer as many queries correctly as possible while keeping Eq.~\eqref{eq:budget}:
\begin{equation}
\max_{B_1,\dots,B_N}\ \sum_{i=1}^{N} \mathbf{1}\big[\text{answer to } q_i \text{ is correct}\big]
\quad \text{s.t. Eq.~\eqref{eq:budget}}.
\end{equation}
A deferred query counts as not correct.
So deferring is never free: it protects the budget, but it lowers accuracy and coverage.
\section{\headlinename Design}
\label{sec:design}
\subsection{Overview}
\sys has two modes that share the same budget ladder, GPU monitor (NVML~\cite{nvml}), and energy accounting.
They differ in \emph{when} they decide and \emph{what they learn}.
The default mode, \atagt, decides step by step, using the samples it has already generated and a cost model that it learns while it runs.
The offline mode, \atcal, decides the whole budget once, before generating any sample, using models built offline.
Table~\ref{tab:design_comparison} summarizes the differences.

\begin{table}[htpb]
\centering
\caption{The two \sys modes.}
\label{tab:design_comparison}
\footnotesize
\begin{tabular}{@{}lll@{}}
\toprule
\textbf{Property} & \textbf{\atcal} & \textbf{\atagt} \\
\midrule
Value of more samples   & Offline accuracy model & Live vote stability \\
Cost model              & Built offline, fixed   & Learned online (RLS) \\
Uses live GPU state     & Yes                    & Yes \\
Uses current samples    & No                     & Yes \\
When $B$ is chosen      & Once, before sampling  & Step by step \\
Dataset profiling       & Required               & Not required \\
Decision score          & $\Delta\hat{P}/\Delta\hat{E}$ & $\Delta U/E_{\mathrm{inc}}$ \\
\bottomrule
\end{tabular}
\end{table}
\noindent\textbf{When to use the offline mode.}
\atagt is the right choice when the load changes or new models and datasets arrive often.
\atcal is useful in three cases: (1)~the model, GPU, and workload stay the same long enough to reuse one profile, where it reaches higher accuracy under stable load (\S\ref{sec:calibration_strength}); (2)~the service needs low tail latency or high throughput on short tasks, since it generates all samples in one batch (\S\ref{sec:overhead}); and (3)~the cost of each query must be known before it runs.

\subsection{\atagt}
\label{sec:agentic}

\atagt removes the offline phase.
It uses two kinds of feedback while it runs: the answers it has already generated for the current query, and the measured cost of the steps it has already executed.
Algorithm~\ref{alg:agentic} shows the loop.

\noindent\textbf{Step-by-step sampling.}
\atagt starts with $B=1$.
It then climbs the same ladder $1\!\to\!3\!\to\!4\!\to\!6\!\to\!8$, but it only generates the \emph{new} samples at each step.
Samples from earlier steps are kept and reused.

\noindent\textbf{Reasoning signal: vote stability.}
After each step, \atagt groups the answers and counts votes.
Let $w_B$ be the current plurality answer after $B$ samples.
The \emph{stability} is the chance that $w_B$ would still win if we went all the way to $B_{\max}=8$:
\begin{equation}
S(B)=P\big(w_B \text{ is still the plurality answer at } B_{\max}\big).
\end{equation}
\atagt estimates $S$ with a small Monte Carlo simulation of the remaining samples, based on the current vote counts.
The expected gain of the next step is the increase in stability:
\begin{equation}
\Delta U = S_{\text{after}} - S_{\text{now}}.
\label{eq:agentic_gain}
\end{equation}
This signal does not need the correct answer.
It only asks: \emph{would more samples still change the result?}

\noindent\textbf{Resource signal: online cost learning.}
After each executed step, \atagt records the measured energy $e_t$ of that step and a feature vector $\boldsymbol{\phi}_t$ that describes the step (the number of new samples) and the current GPU state.
It predicts energy with a linear model, $\hat{E}=\boldsymbol{\theta}^{\top}\boldsymbol{\phi}$, and updates $\boldsymbol{\theta}$ with the standard recursive least squares (RLS) rule~\cite{haykin2002adaptive}:
\begin{align}
\mathbf{k}_t &= \frac{\mathbf{P}_{t-1}\boldsymbol{\phi}_t}{\lambda+\boldsymbol{\phi}_t^{\top}\mathbf{P}_{t-1}\boldsymbol{\phi}_t}, \\
\boldsymbol{\theta}_t &= \boldsymbol{\theta}_{t-1}+\mathbf{k}_t\big(e_t-\boldsymbol{\theta}_{t-1}^{\top}\boldsymbol{\phi}_t\big), \\
\mathbf{P}_t &= \tfrac{1}{\lambda}\big(\mathbf{P}_{t-1}-\mathbf{k}_t\boldsymbol{\phi}_t^{\top}\mathbf{P}_{t-1}\big),
\end{align}
where $\lambda\le 1$ is a forgetting factor ($\lambda=1$ means no forgetting).
RLS is cheap (a few small matrix operations per step), needs no training data, and starts learning from the first query.
Because measurements arrive after every step, the model quickly sees the effect of a load change.
A second RLS model predicts step latency in the same way.

\noindent\textbf{Safety margin.}
To lower the risk of underestimating the cost, \atagt uses a conservative estimate for the next step:
\begin{equation}
E_{\mathrm{inc}} = \hat{E} + \beta\,\sigma_E,
\label{eq:safe_energy}
\end{equation}
where $\sigma_E$ is the uncertainty of the current prediction and $\beta$ sets how careful the controller is.

\noindent\textbf{Continue or stop.}
\atagt continues only if the next step is worth its cost and fits in the budget:
\begin{equation}
\frac{\Delta U}{E_{\mathrm{inc}}} \ge \gamma
\quad\text{and}\quad
E_{\mathrm{inc}} \le E_{\mathrm{rem}},
\label{eq:agentic_score}
\end{equation}
where $\gamma$ is a minimum gain per joule and $E_{\mathrm{rem}}$ is the remaining energy budget.
It also stops when the answer is already stable, when more samples are unlikely to resolve a disagreement, or when it reaches $B_{\max}$.

\begin{algorithm}[t]
\caption{\atagt for one query $q$}
\label{alg:agentic}
\small
\begin{algorithmic}[1]
\Require query $q$; ladder $1\!\to\!3\!\to\!4\!\to\!6\!\to\!8$; remaining budget $E_{\mathrm{rem}}$; RLS model $(\boldsymbol{\theta},\mathbf{P})$
\State $H \gets$ read GPU state (NVML)
\If{$E_{\mathrm{inc}}(B{=}1,H) > E_{\mathrm{rem}}/N_{\mathrm{rem}}$} \Return \textsc{Defer}
\EndIf
\State generate 1 sample; $B\gets 1$; update votes $v$
\State measure step energy $e$; RLS update; $E_{\mathrm{rem}}\gets E_{\mathrm{rem}}-e$
\While{$B<8$ \textbf{and} answer not yet stable}
    \State $B' \gets$ next budget on the ladder
    \State $\Delta U \gets$ \Call{MonteCarloGain}{$v, B\!\to\!B'$} \Comment{Eq.~\eqref{eq:agentic_gain}}
    \State $H \gets$ read GPU state; $E_{\mathrm{inc}} \gets \hat{E}(B'{-}B, H)+\beta\sigma_E$
    \If{$\Delta U/E_{\mathrm{inc}}<\gamma$ \textbf{or} $E_{\mathrm{inc}}>E_{\mathrm{rem}}$} \textbf{break}
    \EndIf
    \State generate $B'{-}B$ new samples; update $v$; $B\gets B'$
    \State measure $e$; RLS update $(\boldsymbol{\theta},\mathbf{P})$; $E_{\mathrm{rem}}\gets E_{\mathrm{rem}}-e$
\EndWhile
\State \Return plurality answer of $v$
\end{algorithmic}
\end{algorithm}

\noindent\textbf{A simple example.}
The following case is only an illustration of the rule.
Suppose a math question has three samples, and all three give the answer 42.
The vote is already stable, so $\Delta U$ is close to zero and \atagt stops at $B=3$, whether the GPU is idle or busy.
Now suppose the three samples give 42, 42, and 40.
More samples could still change the winner, so $\Delta U$ is large.
If the GPU is idle, the next step is cheap, $\Delta U/E_{\mathrm{inc}}$ passes the threshold, and \atagt goes to $B=4$.
If the GPU is busy, the same step costs several times more joules (Fig.~\ref{fig:motivation}(a)), the ratio can drop below $\gamma$, and \atagt returns 42.
In both cases the decision uses the same two questions: is the next sample useful, and what does it cost now?

\noindent\textbf{Two feedback loops.}
Each executed step feeds back twice.
The new answers update the reasoning signal $\Delta U$, and the measured energy updates the cost model.
So the next decision depends on both how the reasoning is going and how the GPU is behaving now.

\noindent\textbf{Overhead.}
The controller runs on the host CPU.
Per step, it reads NVML, runs a small Monte Carlo estimate, and does one RLS update.
The main runtime cost is not this computation, but the fact that samples are generated in several smaller steps instead of one batch (\S\ref{sec:overhead}).

\subsection{\atcal}
\label{sec:calibrated}
\atcal is the offline mode of \sys. We call it \emph{Calibrated} because its accuracy and cost models come from an offline calibration (profiling) phase; it still serves queries at runtime and reads the live GPU state, but its models stay fixed.
For each dataset, we run 500 calibration prompts at each of the five budgets and four load levels (20 settings).
Since the five budgets together use $1+3+4+6+8=22$ samples per prompt, this generates $500\times22\times4=44{,}000$ samples per dataset.
The calibration prompts come from a separate split and do not overlap with the evaluation queries.
From these runs we fit two models.
The \emph{accuracy model} $\hat{P}(q,B)$ estimates the chance that budget $B$ gives a correct answer for query $q$, using query features such as estimated difficulty.
The \emph{cost model} $\hat{E}(B,H)$ predicts the energy of budget $B$ under GPU condition $H$ by predicting its power and run time and multiplying them, since energy is power times time (watts $\times$ seconds $=$ joules).
Both models stay fixed after this phase.

\noindent\textbf{Online phase.}
For each query, \atcal reads the current GPU state $H_t$.
For an upgrade from budget $B$ to the next budget $B'$, it computes the expected accuracy gain and the extra energy:
\begin{align}
\Delta\hat{P}&=\hat{P}(q,B')-\hat{P}(q,B),\\
\Delta\hat{E}&=\hat{E}(B',H_t)-\hat{E}(B,H_t).
\end{align}
The score $\Delta\hat{P}/\Delta\hat{E}$ is the expected gain per extra joule.
The allocator first gives the query the smallest budget.
Then, among the upgrades on the ladder $1\!\to\!3\!\to\!4\!\to\!6\!\to\!8$ that still fit in the remaining energy, it takes the one with the best score.
It stops when no useful upgrade fits.
If even $B=1$ does not fit according to the cost model, the query is deferred.
After this, all $B$ samples are generated and the plurality answer is returned.
The generated samples do not change the chosen budget.

\noindent\textbf{Weakness.}
\atcal uses the live GPU state, but its cost model was learned from the profiling runs and is never updated.
When the real cost moves away from the model, the allocator either overspends early or rejects queries that it could have served.

\section{Experimental Methodology}
\label{sec:methodology}

We evaluate \sys under controlled GPU load.
Our goal is to measure both answer quality and system behavior as the GPU condition changes.

\subsection{Model, Hardware, and Settings}

We use \texttt{Phi-3-mini-4k-instruct}~\cite{phi3mini} in bfloat16 on one NVIDIA A100-SXM4-80GB GPU with a 500~W power limit.
We use tensor parallel size 1 and batch size 32.
Sampling uses temperature 0.7, top-$p$ 0.95, at most 512 new tokens, and a fixed seed (42).
All methods use the same model, prompts, generation settings, and budget set $\mathcal{B}=\{1,3,4,6,8\}$. 
To test generalization (\S\ref{sec:generalization}), we repeat the study on a second GPU type, an NVIDIA L40S with a 350~W power limit, and with a second model, \texttt{Qwen2.5-3B-Instruct}~\cite{qwen25}, using the same precision (bfloat16), batch size, sampling settings, query sets, and load schedule.

\subsection{Datasets}

We use four datasets with different answer lengths: two math datasets with long answers, GSM8K~\cite{cobbe2021gsm8k} and MATH~\cite{hendrycks2021math}, and two multiple-choice datasets with short answers, ARC-Challenge~\cite{clark2018arc} and CommonsenseQA~\cite{talmor2019csqa}.
Each dataset has a development set and a test set of 500 queries each, which do not overlap; all reported results use the test sets. We report accuracy. All query sets and controller settings were frozen before we looked at the results, and we did no tuning afterwards.

\subsection{Compared Methods}

\begin{itemize}
    \item \textbf{\fixed:} standard self-consistency~\cite{wang2023selfconsistency} with a fixed budget: every answered query gets $B=4$ samples, and the plurality answer is returned. We use $B=4$ because it is the reference budget that defines $E_{\max}$.
    \item \textbf{\adcal:} adaptive budget allocation with offline calibration~\cite{kashaniyan2026adaptive}. It chooses $B$ per query from offline accuracy and cost models, but does not use the live GPU state.
    \item \textbf{\atcal} (\S\ref{sec:calibrated}): the offline mode of \sys.
    \item \textbf{\atagt} (\S\ref{sec:agentic}): the default mode of \sys.
\end{itemize}

\noindent\textbf{Choice of baselines.}
Fixed-budget self-consistency is the base method that all self-consistency methods in Table~\ref{tab:related} build on, and \adcal is the closest prior work, as the only one that models energy.
The other adaptive methods stop sampling from reasoning signals alone, so their budget cannot react to GPU load; our ReasoningOnly-Strict ablation, which uses only such a signal, keeps the same budget at every load level (\S\ref{sec:causal_ablation}).

\subsection{Controlled GPU Load}
To create a shared-GPU setting, we run a background load injector on the same GPU. It competes with the LLM for GPU resources, like other jobs on a shared GPU. The injector is a background thread that runs dense FP16 matrix multiplies on the same GPU for a duty fraction of every 0.2~s period and sleeps for the rest.
We use four levels: \emph{idle} (no injector), \emph{low}, \emph{medium}, and \emph{high}, with injector duty settings of 0.25, 0.55, and 0.85. The duty setting controls how hard the injector works; it is not a utilization target.
We check the real effect with NVML~\cite{nvml}.
For example, in the \fixed runs the measured GPU utilization is about 0\%, 24\%, 54\%, and 85\%, and the average GPU power on GSM8K rises from 77~W at idle to 431~W at high load. The 500 queries of each run are split equally: 125 queries per load level, in the order idle, low, medium, high.
The schedule is fixed by a hash and is reused exactly for every method.
So all methods see the same queries in the same order under the same injector settings.

\begin{table*}[t]
\centering
\caption{Overall results under the idle$\to$low$\to$medium$\to$high load schedule (500 queries per dataset). Our default mode, \atagt, is shaded. Accuracy counts deferred queries as wrong. Mean $B$ counts deferred queries as $B=0$. Budget used above 100\% (red) means the energy budget was exceeded. Bold: best value per dataset.}
\label{tab:overall_results}
\footnotesize
\setlength{\tabcolsep}{3.6pt}
\begin{minipage}[t]{0.495\textwidth}
\centering
\begin{tabular}{@{}llrrrrr@{}}
\toprule
& \textbf{Method} & \textbf{Acc.}$\uparrow$ & \textbf{Defer}$\downarrow$ & \textbf{Mean} & \textbf{J/corr.}$\downarrow$ & \textbf{Budget} \\
&  &  & \textbf{(\%)} & \textbf{$B$} &  & \textbf{used (\%)} \\
\midrule
\multirow{4}{*}{\rotatebox{90}{GSM8K}}
& \fixed & 0.364 & 51.8 & 1.928 & 312.4 & 93.1 \\
& \adcal & 0.368 & 48.8 & 1.726 & 330.5 & 99.5 \\
& \atcal & 0.600 & 11.2 & 1.796 & 196.5 & 96.5 \\
& \ours\atagt & \ours\textbf{0.626} & \ours\textbf{0.0} & \ours1.580 & \ours\textbf{191.9} & \ours98.3 \\
\midrule
\multirow{4}{*}{\rotatebox{90}{MATH}}
& \fixed & 0.246 & 49.6 & 2.016 & 600.5 & 86.1 \\
& \adcal & 0.200 & 55.2 & 1.474 & 790.7 & 92.2 \\
& \atcal & 0.424 & 6.6 & 2.046 & 401.2 & 99.2 \\
& \ours\atagt & \ours\textbf{0.426} & \ours\textbf{0.0} & \ours1.794 & \ours\textbf{371.2} & \ours92.2 \\
\bottomrule
\end{tabular}
\end{minipage}\hfill
\begin{minipage}[t]{0.495\textwidth}
\centering
\begin{tabular}{@{}llrrrrr@{}}
\toprule
& \textbf{Method} & \textbf{Acc.}$\uparrow$ & \textbf{Defer}$\downarrow$ & \textbf{Mean} & \textbf{J/corr.}$\downarrow$ & \textbf{Budget} \\
&  &  & \textbf{(\%)} & \textbf{$B$} &  & \textbf{used (\%)} \\
\midrule
\multirow{4}{*}{\rotatebox{90}{ARC-C}}
& \fixed & 0.568 & 30.8 & 2.768 & 25.1 & 87.4 \\
& \adcal & 0.620 & 23.2 & 2.450 & 28.0 & \textcolor{tabred}{106.4} \\
& \atcal & 0.722 & 9.0 & 2.878 & \textbf{19.2} & 84.9 \\
& \ours\atagt & \ours\textbf{0.794} & \ours\textbf{0.0} & \ours1.798 & \ours19.4 & \ours94.7 \\
\midrule
\multirow{4}{*}{\rotatebox{90}{CSQA}}
& \fixed & 0.666 & 8.6 & 3.656 & 17.8 & \textcolor{tabred}{102.3} \\
& \adcal & 0.688 & 5.8 & 3.642 & 16.7 & 99.1 \\
& \atcal & 0.686 & 4.6 & 4.192 & \textbf{13.3} & 78.7 \\
& \ours\atagt & \ours\textbf{0.716} & \ours\textbf{0.0} & \ours2.848 & \ours16.2 & \ours99.9 \\
\bottomrule
\end{tabular}
\end{minipage}
\end{table*}

\subsection{Energy Accounting and Budget}

We measure energy by GPU occupancy ($\Eocc$). Each microbatch is charged for all the GPU energy used while it runs, computed from NVML power readings and its run time.
When other jobs share the GPU, a microbatch runs longer and the GPU draws more power, so its $\Eocc$ grows.
We use this accounting because a job on a shared GPU holds the device for its whole run time, so this is the energy it is responsible for; the same accounting is used for all methods, so it does not favor any of them.
Each run gets one shared energy budget $E_{\max}$, set from the reference energy needed to run $B=4$ under the reference GPU condition. We set this budget separately for each GPU, model, and dataset, because the cost of a sample depends on all three.
All methods get the same budget.

\subsection{Metrics}

\noindent\textbf{Quality and coverage.}
\emph{Accuracy} is computed over \emph{all} 500 queries; a deferred query counts as wrong.
\emph{Defer} is the percentage of deferred queries.

\noindent\textbf{Sampling effort.}
\emph{Mean $B$} is the average budget.
In overall results we count a deferred query as $B=0$; in per-load results we average only over answered queries, so that the budget choice is visible.

\noindent\textbf{Energy.}
We report $\Eocc$ per query and \emph{joules per correct answer}:
\begin{equation}
\text{J/correct}=\frac{\text{total } \Eocc}{\text{number of correct answers}}.
\end{equation}
This metric gets worse both when a method spends energy on samples that do not help and when it gets fewer correct answers, for example by deferring queries.

\noindent\textbf{Budget safety.}
\emph{Budget used} is total energy divided by $E_{\max}$.
A value above 100\% means the method \emph{overshot} the budget.

\noindent\textbf{System performance.}
We report 95th-percentile (p95) latency per query, query throughput (queries per second, QPS), GPU utilization, and power.
For \atagt we also report the error of its online cost model as mean absolute percentage error (MAPE).

\subsection{Calibration Cost}

We measure the offline profiling of the calibrated methods separately, because it is paid before serving starts.
Over four datasets, it uses 2,000 prompts, 80 settings, and 176,000 generated samples, for a total of 5.73 GPU-hours and 7.16~MJ.
\atagt needs no dataset-specific profiling.

\subsection{Controlled Ablation}
\label{sec:ablation_setup}

In the main runs, the remaining budget shrinks over time, so a change in $B$ could come from the budget and not from the load.
To separate these effects, we run three versions of \atagt:
\textbf{Full-HW} (live GPU state and online cost learning),
\textbf{noHW-State} (online cost learning, but the GPU-state input is hidden), and
\textbf{ReasoningOnly-Strict} (no GPU state and online cost learning frozen).
In this experiment we use a large, non-binding energy budget and keep the remaining-energy value that the controller sees fixed.
So the budget cannot drive the decision.

\section{Evaluation}
\label{sec:evaluation}

We organize the results around seven questions:
\textbf{Q1:} Does \atagt keep quality and serve every query under changing load?
\textbf{Q2:} How does each controller react as load rises, and why does the offline mode defer?
\textbf{Q3:} What does each mode cost in energy, including profiling?
\textbf{Q4:} Is the adaptation really caused by the resource signals?
\textbf{Q5:} What is the runtime overhead of online control?
\textbf{Q6:} When is offline calibration the better choice?
\textbf{Q7:} Does the behavior hold on another GPU and another model?

\begin{figure*}[t]
    \centering
    \includegraphics[width=\textwidth]{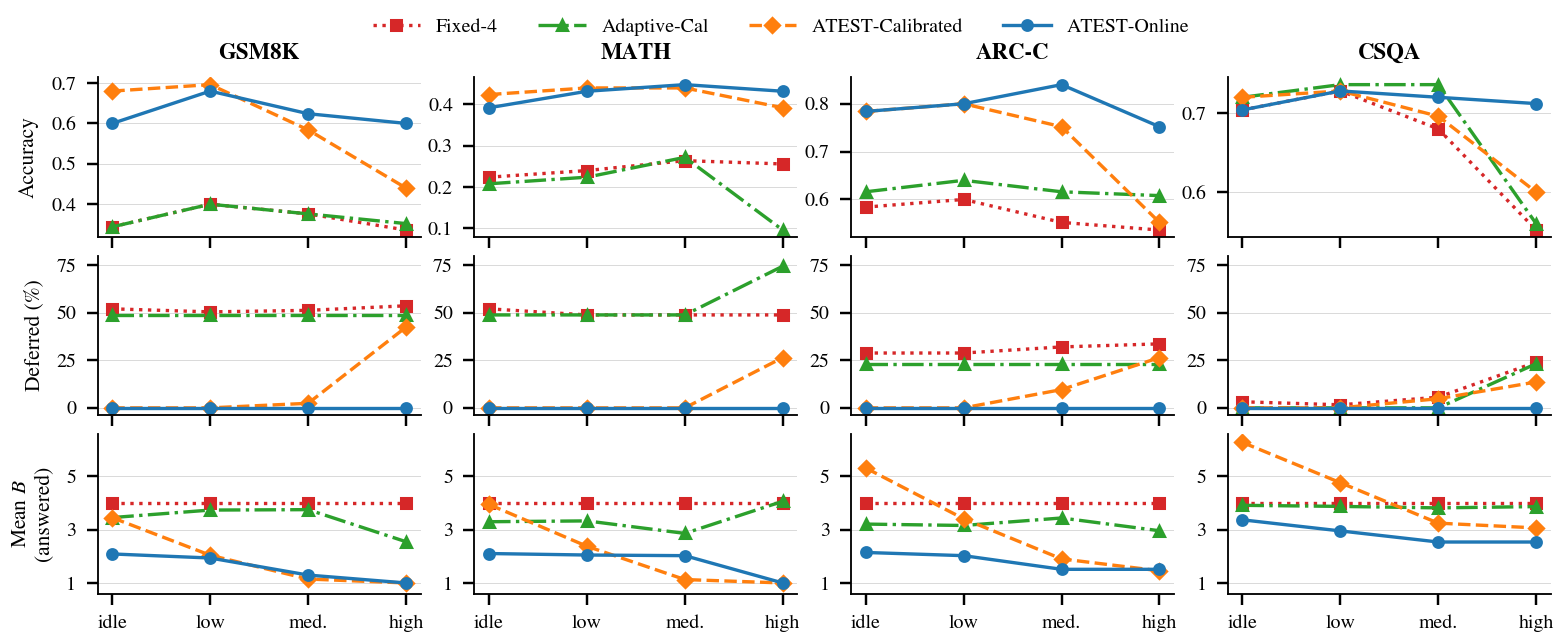}
    \caption{Behavior at each load level (125 queries per level). Top: accuracy over all queries (deferred = wrong). Middle: deferred queries (\%). Bottom: mean $B$ over \emph{answered} queries. \atagt (blue, solid) never defers and lowers its budget as load rises. \atcal (orange, dashed) also lowers its budget, but it starts deferring at medium or high load on all four datasets, which pulls its accuracy down.}
    \label{fig:per_load}
\end{figure*}

\subsection{Q1: Overall Results under Changing Load}
\label{sec:overall_results}

Table~\ref{tab:overall_results} gives the main results.
Three points stand out.

\noindent\textbf{\atagt serves every query.}
It has zero deferred queries on all four datasets and stays within the energy budget on all of them (budget used 92--100\%).
Every other method defers queries on all four datasets, and \fixed and \adcal also exceed the budget once each.

\noindent\textbf{\atagt gives the best or tied-best accuracy.}
It has the highest accuracy on all four datasets: GSM8K (0.626), MATH (0.426), ARC-Challenge (0.794), and CommonsenseQA (0.716).
Each result comes from one run of 500 queries, so the 95\% confidence interval of one accuracy is about $\pm$4 points; gaps of up to about 3 points, as on GSM8K, MATH, and CommonsenseQA, are ties.
The clear gains are on ARC-Challenge (+7.2 points over \atcal), where \atcal loses accuracy mainly because it defers 9.0\% of the queries, and at high load (\S\ref{sec:contention_results}).

\noindent\textbf{\atagt uses fewer samples.}
It has the lowest mean $B$ on three of four datasets, even though it answers more queries than any other method, because it stops as soon as the vote is stable.
On MATH only \adcal is lower, and only because it defers 55.2\% of the queries. The two baselines do much worse on the long-answer datasets: \fixed and \adcal defer about half of the GSM8K and MATH queries, at every load level (Fig.~\ref{fig:per_load}).
Since each query may use only its share of the remaining budget, four samples often do not fit, so these baselines defer queries from the start.

\noindent\textbf{Answer to Q1:} Yes. Under changing load, \atagt answers every query, stays within budget, and gives the best or tied-best accuracy on all four datasets.

\subsection{Q2: Behavior as the Load Rises}
\label{sec:contention_results}

Fig.~\ref{fig:per_load} shows each method at each load level.

\noindent\textbf{Both \sys modes use fewer samples when the GPU is busy.}
\atagt lowers its mean budget from idle to high load on all four datasets: from 2.09 to 1.00 on GSM8K, 2.10 to 1.00 on MATH, 2.14 to 1.51 on ARC-Challenge, and 3.37 to 2.54 on CommonsenseQA.
\atcal also lowers its budget (for example, from 3.45 to 1.00 on GSM8K).
\fixed always uses $B=4$ for the queries it answers.

\noindent\textbf{\atagt keeps serving at high load.}
On GSM8K at high load, both \sys modes end up at $B=1$.
But \atcal defers 53 of the 125 queries (42.4\%), so its accuracy for this phase falls to 44.0\%.
\atagt answers all 125 queries with 60.0\% accuracy and uses less energy per query (175.8~J vs.\ 206.4~J).
On MATH at high load the pattern is the same: \atcal defers 26.4\% and reaches 39.2\%, while \atagt defers none and reaches 43.2\%.

\noindent\textbf{Why does \atcal defer?}
We see two reasons in the measurements.
\emph{First, it spends more early.}
At idle on GSM8K, \atcal uses $B=3.45$ and 124.4~J per query, while \atagt uses $B=2.09$ and 60.9~J.
This buys higher idle accuracy (68.0\% vs.\ 60.0\%), but the budget is shared by the whole run, so less is left for the high-load phase, when each query is much more expensive.
\emph{Second, its fixed cost model can be too pessimistic.}
\atcal admits a query only if the predicted cost fits.
On CommonsenseQA it deferred 23 queries although 21\% of the energy budget was never used; on ARC-Challenge it deferred 45 queries with 15\% unused.
So the budget was not the limit here; the deferrals came from the controller's own cost predictions.
Fig.~\ref{fig:motivation}(b) shows the root cause: the offline cost model is often far from the real cost under load.
\atagt avoids both problems because it measures every step and corrects its cost model as it goes.

\noindent\textbf{A note on measured load.}
The injector settings are identical for all methods, but GPU utilization is sampled before each microbatch, so the reading depends on each method's execution pattern.
On GSM8K and MATH, all methods see the same high-load utilization (about 85\%); for \atagt on ARC-Challenge and CommonsenseQA the sampled reading at high load is lower (68\% and 45\%).
We therefore base our main high-load comparison on GSM8K and MATH.

\noindent\textbf{Answer to Q2:} As load rises, \atagt lowers its budget and keeps serving every query. \atcal also lowers its budget, but its early spending and fixed cost model lead to deferrals at high load.

\subsection{Q3: Energy Cost, Including Profiling}
\label{sec:energy_results}

\noindent\textbf{Serving energy.}
\atagt has the lowest joules per correct answer on GSM8K (191.9~J) and MATH (371.2~J), the two long-answer datasets where the energy budget is tight.
On ARC-Challenge the two modes are almost equal (19.4~J vs.\ 19.2~J).
On CommonsenseQA, \atcal is more efficient (13.3~J vs.\ 16.2~J, Table~\ref{tab:overall_results}), but it leaves 4.6\% of the queries unanswered.
So \atagt is not always the cheapest per correct answer during serving, but it wins the overall trade-off: it gives higher accuracy, answers every query, and, once profiling is counted, uses far less total energy on every dataset.

\begin{table}[t]
\centering
\caption{End-to-end cost of The two \sys modes on four datasets (2,000 queries under the load schedule). \atcal must profile every dataset before serving. \atagt starts serving at once and learns its cost model during serving, so its learning cost is included in the serving numbers.}
\label{tab:setup_vs_serving}
\footnotesize
\begin{tabular}{@{}lrr@{}}
\toprule
& \textbf{\atcal} & \textbf{\atagt} \\
\midrule
\multicolumn{3}{@{}l}{\emph{Offline profiling (before serving)}} \\
GPU time (h)        & 5.73    & \ours 0 \\
Energy (MJ)         & 7.16    & \ours 0 \\
Generated samples   & 176,000 & \ours 0 \\
\midrule
\multicolumn{3}{@{}l}{\emph{Serving 2,000 queries}} \\
Run time (h)        & 0.14    & \ours 0.14 \\
Energy (MJ)         & 0.156   & \ours 0.153 \\
Queries answered    & 1,843   & \ours \textbf{2,000} \\
Correct answers     & 1,216   & \ours \textbf{1,281} \\
\midrule
\multicolumn{3}{@{}l}{\emph{Total}} \\
GPU time (h)        & 5.87    & \ours \textbf{0.14} \\
Energy (MJ)         & 7.31    & \ours \textbf{0.153} \\
\bottomrule
\end{tabular}
\end{table}

\noindent\textbf{Setup cost versus serving cost.}
Table~\ref{tab:setup_vs_serving} puts the full cost of both modes side by side.
\atcal must profile every dataset before it can serve a single query: 5.73 GPU-hours, 7.16~MJ, and 176,000 generated samples for our four datasets.
\atagt skips this stage completely and starts serving at once; its cost model learns during normal serving, so its learning cost is already inside the serving numbers.
Serving itself costs almost the same for both modes (about 0.14 hours and 0.15~MJ for 2,000 queries), but \atagt answers all 2,000 queries and gives 65 more correct answers.
In total, \atcal uses about 42$\times$ more GPU time and 48$\times$ more energy than \atagt in our study.
Calibration is a one-time cost per dataset, so a long, stable deployment can spread it over many queries, but it must be repeated whenever the model, the GPU, the prompt format, or the workload changes.

\noindent\textbf{Answer to Q3:} \atagt is the most energy-efficient mode on the long-answer datasets. Including profiling, it uses 42$\times$ less GPU time and 48$\times$ less energy than \atcal for the same 2,000 queries.

\subsection{Q4: Is the Adaptation Caused by Resource Signals?}
\label{sec:causal_ablation}

\begin{table}[t]
\centering
\caption{Controlled ablation of \atagt with a non-binding budget and a fixed remaining-energy signal (\S\ref{sec:ablation_setup}). ``Load response'' means that mean $B$ changes across load levels. ReasoningOnly-Strict uses the same mean budget, $B=3.12$ samples per query, at every load level.}
\label{tab:resource_ablation}
\footnotesize
\begin{tabular}{@{}lccl@{}}
\toprule
\textbf{Variant} & \textbf{GPU state} & \textbf{Online learning} & \textbf{Load response} \\
\midrule
Full-HW              & Yes & Yes & Yes \\
noHW-State           & No  & Yes & Yes \\
ReasoningOnly-Strict & No  & No  & No ($B{=}3.12$) \\
\bottomrule
\end{tabular}
\end{table}

In the main runs, the energy budget shrinks over time.
So the end-to-end results alone cannot tell us whether \atagt lowers $B$ because of the load or because the budget is running out.
The controlled ablation (\S\ref{sec:ablation_setup}) removes this confound: the budget is large enough never to bind, and the remaining-energy value seen by the controller is fixed.

Table~\ref{tab:resource_ablation} shows the result.
When both the GPU state and online cost learning are removed (ReasoningOnly-Strict), the controller uses the same mean budget, 3.12 samples per query at every load level on all four datasets.
Its reasoning signal still changes from query to query, but without resource feedback its decisions do not react to the load at all.
When online cost learning is on (noHW-State), the budget changes with the load even though the GPU-state input is hidden.
This happens because the measured energy of each step already reflects the load.
Adding the GPU state (Full-HW) changes the budgets again, most visibly on GSM8K and MATH.

Resource awareness does not mean that $B$ must always go down when load goes up.
The decision also depends on how stable the vote is.
What the ablation shows is that removing the resource signals removes the controller's response to the load.

\noindent\textbf{Answer to Q4:} Yes. With budget effects removed, online cost learning is necessary for the budget to follow the load, and the GPU state further changes the decisions.

\subsection{Q5: Runtime Overhead and Cost-Model Error}
\label{sec:overhead}
\label{sec:system_performance}

\begin{table}[t]
\centering
\caption{Runtime behavior of the two \sys modes under the load schedule: p95 latency per query (s), throughput (QPS), and the energy error of the online cost model of \atagt (MAPE).}
\label{tab:system_performance}
\footnotesize
\setlength{\tabcolsep}{4pt}
\begin{tabular}{@{}lrrrrr@{}}
\toprule
& \multicolumn{2}{c}{\textbf{\atcal}} & \multicolumn{3}{c}{\textbf{\atagt}} \\
\cmidrule(lr){2-3}\cmidrule(l){4-6}
\textbf{Dataset} & \textbf{p95} & \textbf{QPS} & \textbf{p95} & \textbf{QPS} & \textbf{MAPE} \\
\midrule
GSM8K    & 0.509 & 2.62  & 1.081 & 2.62  & 23\% \\
MATH     & 1.019 & 1.78  & 1.608 & 1.99  & 22\% \\
ARC-C    & 0.130 & 19.72 & 0.111 & 17.19 & 134\% \\
CSQA     & 0.086 & 29.69 & 0.180 & 16.60 & 112\% \\
\bottomrule
\end{tabular}
\end{table}

\noindent\textbf{Latency and throughput.}
Online control is not free (Table~\ref{tab:system_performance}).
\atagt has higher p95 latency on three of four datasets, for example 1.08~s vs.\ 0.51~s on GSM8K and 0.18~s vs.\ 0.09~s on CommonsenseQA.
Its throughput is about the same on GSM8K, higher on MATH, and lower on the two multiple-choice datasets, most clearly on CommonsenseQA (16.6 vs.\ 29.7 QPS).
The reason is the step-by-step execution: \atagt may run several small microbatches with a decision between them, instead of one batch with all samples.
For short answers, the fixed cost of each step is large compared with the generation time.

\noindent\textbf{Online cost-model error.}
The online cost model is updated after every step (671 to 993 updates per run).
On the long-answer datasets its energy MAPE is moderate: 23\% on GSM8K and 22\% on MATH.
On the short multiple-choice datasets it is much higher (134\% on ARC-Challenge and 112\% on CommonsenseQA).
One likely reason is that a step there uses only a few joules, so small absolute errors become large relative errors.
Even with this error, \atagt never exceeded the energy budget.
Two design choices help here: the safety margin $\beta\sigma_E$ in Eq.~\eqref{eq:safe_energy}, and the small step size, which limits the damage of one wrong prediction to one step.
A better cost model for very short steps is future work.

\noindent\textbf{Answer to Q5:} The price of online control is higher tail latency and lower throughput on short tasks. The online cost model is accurate enough on long tasks and noisy on short ones, but the controller stays budget-safe.

\begin{table*}[t]
\centering
\caption{Results on a second GPU and a second model (500 queries per dataset, same load schedule, budget rule, and frozen controller settings). A100 values are from Table~\ref{tab:overall_results}. $E_{\max}$ is re-measured for each GPU and model, so joules can be compared within a column group but not across groups. Bold: better of the two methods in each group.}
\label{tab:generalization}
\footnotesize
\setlength{\tabcolsep}{4pt}
\begin{tabular}{@{}llrrrrrrrrr@{}}
\toprule
& & \multicolumn{3}{c}{\textbf{Phi-3-mini on A100}} & \multicolumn{3}{c}{\textbf{Phi-3-mini on L40S}} & \multicolumn{3}{c}{\textbf{Qwen2.5-3B on L40S}} \\
\cmidrule(lr){3-5}\cmidrule(lr){6-8}\cmidrule(l){9-11}
\textbf{Dataset} & \textbf{Method} & \textbf{Acc.}$\uparrow$ & \textbf{Defer (\%)}$\downarrow$ & \textbf{J/corr.}$\downarrow$ & \textbf{Acc.}$\uparrow$ & \textbf{Defer (\%)}$\downarrow$ & \textbf{J/corr.}$\downarrow$ & \textbf{Acc.}$\uparrow$ & \textbf{Defer (\%)}$\downarrow$ & \textbf{J/corr.}$\downarrow$ \\
\midrule
\multirow{2}{*}{GSM8K} & \fixed & 0.364 & 51.8 & 312.4 & 0.506 & 34.2 & 386.2 & 0.138 & 48.8 & 1019 \\
 & \ours\atagt & \ours\textbf{0.626} & \ours\textbf{0.0} & \ours\textbf{191.9} & \ours\textbf{0.638} & \ours\textbf{0.0} & \ours\textbf{303.0} & \ours\textbf{0.280} & \ours\textbf{0.0} & \ours\textbf{536} \\
\midrule
\multirow{2}{*}{MATH} & \fixed & 0.246 & 49.6 & 600.5 & 0.380 & 24.4 & 730.3 & 0.242 & 35.4 & 1067 \\
 & \ours\atagt & \ours\textbf{0.426} & \ours\textbf{0.0} & \ours\textbf{371.2} & \ours\textbf{0.448} & \ours\textbf{0.0} & \ours\textbf{614.3} & \ours\textbf{0.348} & \ours\textbf{0.0} & \ours\textbf{700} \\
\midrule
\multirow{2}{*}{ARC-C} & \fixed & 0.568 & 30.8 & 25.1 & 0.672 & 15.4 & 31.4 & 0.380 & 26.0 & 155 \\
 & \ours\atagt & \ours\textbf{0.794} & \ours\textbf{0.0} & \ours\textbf{19.4} & \ours\textbf{0.792} & \ours\textbf{0.0} & \ours\textbf{26.6} & \ours\textbf{0.488} & \ours\textbf{0.0} & \ours\textbf{123} \\
\midrule
\multirow{2}{*}{CSQA} & \fixed & 0.666 & 8.6 & 17.8 & 0.715 & \textbf{0.0} & \textbf{17.7} & 0.414 & 25.8 & 91 \\
 & \ours\atagt & \ours\textbf{0.716} & \ours\textbf{0.0} & \ours\textbf{16.2} & \ours \textbf{0.718} & \ours\textbf{0.0} & \ours17.9 & \ours\textbf{0.564} & \ours\textbf{0.0} & \ours\textbf{69} \\
\bottomrule
\end{tabular}
\end{table*}

\subsection{Q6: When Does Offline Calibration Win?}
\label{sec:calibration_strength}

\begin{table}[t]
\centering
\caption{Results with a stable GPU (no injected load, 500 test queries per dataset). Accuracy counts deferred queries as wrong: \fixed defers 10--50 queries, \atcal 0--11 queries, and \atagt none.}
\label{tab:stable}
\footnotesize
\setlength{\tabcolsep}{3.2pt}
\begin{tabular}{@{}lrrrrrr@{}}
\toprule
& \multicolumn{3}{c}{\textbf{Accuracy}$\uparrow$} & \multicolumn{3}{c}{\textbf{J/correct}$\downarrow$} \\
\cmidrule(lr){2-4}\cmidrule(l){5-7}
\textbf{Dataset} & \textbf{Fixed-4} & \textbf{Cal.} & \textbf{Online} & \textbf{Fixed-4} & \textbf{Cal.} & \textbf{Online} \\
\midrule
GSM8K    & 0.668 & \textbf{0.686} & 0.632 & 176.5 & \textbf{173.3} & 191.1 \\
MATH     & \textbf{0.446} & 0.402 & 0.402 & \textbf{357.0} & 421.9 & 414.8 \\
ARC-C    & 0.732 & 0.792 & \textbf{0.794} & \textbf{12.6} & 18.4 & 17.3 \\
CSQA     & 0.712 & \textbf{0.726} & 0.716 & \textbf{10.5} & 13.4 & 11.2 \\
\bottomrule
\end{tabular}
\end{table}

Offline calibration is a good choice when the GPU load is stable.
In the idle phase of the load schedule, \atcal matches or beats \atagt in accuracy on all four datasets (for example, 68.0\% vs.\ 60.0\% on GSM8K and 42.4\% vs.\ 39.2\% on MATH).
Table~\ref{tab:stable} shows a separate run with no injected load at all.
Here \atcal has higher accuracy than \atagt on GSM8K (0.686 vs.\ 0.632), and they are close on the other datasets.
Even \fixed is strong in this setting: it has the lowest J/correct on three of four datasets.

\noindent\textbf{Answer to Q6:} When the load is stable and the profile matches the runtime, offline calibration, or even a well-chosen fixed budget, is hard to beat. The advantage of \atagt appears when the load changes.

\subsection{Q7: Does the Behavior Hold on Another GPU and Another Model?}
\label{sec:generalization}

We repeat the experiment on a second GPU, an NVIDIA L40S, with Phi-3-mini and Qwen2.5-3B-Instruct on all four datasets.
Everything else follows the A100 protocol; only $E_{\max}$ is re-measured, because the cost of a sample depends on the GPU and the model.
We compare \fixed with \atagt; \atcal and \adcal would first need a new offline profiling of every dataset on this GPU, which is exactly the cost that \atagt avoids.
Table~\ref{tab:generalization} shows the results.

\noindent\textbf{Zero deferrals.}
\atagt answers every query in all eight new settings within the energy budget (92--99.5\% used), while \fixed defers up to 48.8\%.

\noindent\textbf{Accuracy.}
With Phi-3-mini, \atagt reaches almost the same accuracy on the L40S as on the A100 (0.638 vs.\ 0.626 on GSM8K and 0.448 vs.\ 0.426 on MATH) and keeps its gains over \fixed: +13.2 points on GSM8K, +6.8 on MATH, and +12.0 on ARC-Challenge.
With Qwen2.5-3B, it is more accurate than \fixed on every dataset: it doubles GSM8K accuracy (0.138 to 0.280) and raises MATH from 0.242 to 0.348, ARC-Challenge from 0.380 to 0.488, and CommonsenseQA from 0.414 to 0.564.

\noindent\textbf{Energy.}
\atagt lowers joules per correct answer in seven of eight settings.
With Phi-3-mini on the L40S, the drop is 22\% on GSM8K (303.0 vs.\ 386.2~J), 16\% on MATH, and 15\% on ARC-Challenge; with Qwen2.5-3B it is larger, from 21\% on ARC-Challenge to almost half on GSM8K (536 vs.\ 1019~J).

\noindent\textbf{When the budget is loose.}
The one exception is CommonsenseQA with Phi-3-mini on the L40S.
Here the budget is loose enough for \fixed to answer every query with $B=4$, and the two methods are almost equal (0.715 vs.\ 0.718 accuracy, 17.7 vs.\ 17.9~J).
This agrees with Q6: online control helps most when the budget is tight.

\noindent\textbf{Answer to Q7:} Yes. On a second GPU and a second model, \atagt again answers every query within the budget, is at least as accurate as \fixed in all eight settings, and uses less energy per correct answer in seven of them.
\section{Discussion}
\label{sec:discussion}

\noindent\textbf{Which mode should be used?}
If the model, GPU, and workload stay the same and the load is stable, the profile of \atcal can be built once and reused, giving strong accuracy, low tail latency, and high throughput.
If the GPU is shared, the load changes, or new datasets and models arrive often, \atagt is the safer choice: it needs no profiling, follows the live cost, and serves every query.
The two modes can also be combined by starting \atagt from a calibrated cost model.

\noindent\textbf{Service coverage is a systems metric.}
A controller can look good on the queries it answers while refusing many others, so we count deferred queries as wrong.
On GSM8K at high load, this shows the real difference: 44.0\% accuracy with 42.4\% deferrals for \atcal, against 60.0\% with none for \atagt.

\noindent\textbf{Resource awareness is more than ``use fewer samples when busy.''}
\atagt compares the value of the next step with its current cost: it stops on a stable vote even on an idle GPU, and continues on a split vote even under load if the cost is acceptable.

\noindent\textbf{Why not let an LLM decide?}
A variant in which an LLM makes the continue-or-stop decision reached similar accuracy, but exceeded the energy budget on three of four datasets (by up to 6.2\% on GSM8K) and spent extra GPU energy on its decisions.
A small numerical controller is cheaper and easier to keep within a hard budget.
\section{Limitations and Future Work}
\label{sec:limitations}

\noindent\textbf{Evaluation scope.}
Our evaluation is bounded by the available compute.
Every setting must be run for every method and load level, together with the stable-load runs, the ablations, and the offline calibration, so the total GPU time grows quickly.
We therefore use 500 queries per dataset, four datasets, two open models, and two GPU types, which is enough to compare all methods under the same conditions. 

\noindent\textbf{Runtime overhead.}
Step-by-step control adds some overhead on short tasks (\S\ref{sec:overhead}).
Batching the steps of several queries and overlapping decisions with generation could reduce it.

\noindent\textbf{Online cost model.}
The online cost model of \atagt is simple by design.
Richer online models could
make cost predictions more precise.

\noindent\textbf{Integration with inference systems.}
\sys decides how much reasoning each query gets, while inference engines and power managers decide how that work runs.
Running \sys inside engines such as vLLM~\cite{kwon2023vllm} or with GPU frequency control~\cite{kakolyris2025throttle} could manage both together.

\section{Conclusion}
\label{sec:conclusion}

The cost of a test-time scaling sample is not fixed on a shared GPU; in our measurements it changes by up to 4.1$\times$ with load.
We presented \sys, which makes the sampling decision depend on both the value of more reasoning and its current cost.
Its default mode, \atagt, learns the cost online, decides step by step, and needs no profiling; its offline mode, \atcal, works well when the load is stable and one profile can be reused.
Under changing load on four datasets, \atagt answered every query, stayed within the energy budget, and gave the highest or tied-best accuracy, while the offline mode deferred up to 42.4\% of queries at high load. The same behavior held on an L40S GPU and with a second model, Qwen2.5-3B.
It also avoided 5.73 GPU-hours of offline calibration.
A controlled ablation showed that online cost learning is what makes the budget follow the load.
The price is higher tail latency and lower throughput on short tasks.
Our results suggest that test-time scaling on shared accelerators should be treated as a runtime resource-management problem, not only as a reasoning problem.

\balance
\bibliographystyle{IEEEtran}
\bibliography{references}

\end{document}